\documentclass[runningheads]{llncs}

\usepackage[table]{xcolor}
\usepackage{todonotes}
\usepackage{lipsum}
\usepackage{dsfont}
\usepackage{hyperref}
\usepackage{graphicx}
\usepackage{multirow}
\usepackage{multicol}
\usepackage{wrapfig}
\usepackage{subcaption}
\usepackage{booktabs}

\AtBeginDocument{%
  \let\mathbb\relax
  \DeclareMathAlphabet\PazoBB{U}{fplmbb}{m}{n}
  \newcommand{\mathbb}{\PazoBB}
}

\begin{document}
\title{Context-driven Visual Object Recognition based on Knowledge Graphs}

%
%
\author{Sebastian Monka\inst{1,2} \and
Lavdim Halilaj\inst{1} \and
Achim Rettinger\inst{2}}
\authorrunning{S. Monka et al.}
\titlerunning{}



%
\institute{Bosch Center for Artificial Intelligence, Renningen, Germany \\
\email{\{sebastian.monka,lavdim.halilaj\}@de.bosch.com} \and
Trier University, Trier, Germany \\
\email{\{rettinger\}@uni-trier.de}}
%

\maketitle              

\begin{abstract}

Current deep learning methods for object recognition are purely data-driven and require a large number of training samples to achieve good results.
Due to their sole dependence on image data, these methods tend to fail when confronted with new environments where even small deviations occur.
Human perception, however, has proven to be significantly more robust to such distribution shifts.
It is assumed that their ability to deal with unknown scenarios is based on extensive incorporation of contextual knowledge.
Context can be based either on object co-occurrences in a scene or on memory of experience.
In accordance with the human visual cortex which uses context to form different object representations for a seen image, we propose an approach that enhances deep learning methods by using external contextual knowledge encoded in a knowledge graph.
Therefore, we extract different contextual views from a generic knowledge graph, transform the views into vector space and infuse it into a DNN.
We conduct a series of experiments to investigate the impact of different contextual views on the learned object representations for the same image dataset.
The experimental results provide evidence that the contextual views influence the image representations in the DNN differently and therefore lead to different predictions for the same images.
We also show that context helps to strengthen the robustness of object recognition models for out-of-distribution images, usually occurring in transfer learning tasks or real-world scenarios.

\end{abstract}
\section{Introduction}
\begin{figure}[tb]
    \centering
    \begin{subfigure}[b]{0.49\textwidth}
        \centering
        \includegraphics[height=0.4\textwidth]{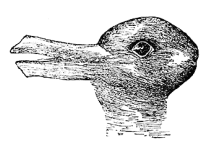}
        \caption{Duck or rabbit?~\cite{Jastrow1900}}
        \label{fig:duck_rabbit}
    \end{subfigure}
    \hfill
    \begin{subfigure}[b]{0.49\textwidth}
        \centering
        \includegraphics[height=0.4\textwidth]{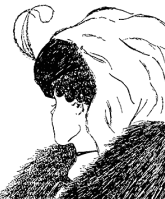}
        \caption{Young lady or old woman?~\cite{Attneave1971}.}
        \label{fig:young_old_women}
    \end{subfigure}
    \caption{The mental representation for ambiguous images can change based on the context, although the perceived image is still the same.}
    \label{fig:optical_illusions}
\end{figure}

How humans perceive the real world is strongly dependent on the context~\cite{OLIVA2007520,Lauer2021}.
Especially, in situations with poor quality of visual input, for instance caused by large distances, or short capturing times, context appears to play a major role in improving the reliability of recognition~\cite{DBLP:journals/ijcv/Torralba03}.
Perception is not only influenced by co-occurring objects or visual features in the same image, but also by experience and memory~\cite{Rafetseder2021}.
There is evidence that humans perceive similar images differently considering the given context~\cite{Chambers1985}.
A famous example are ambiguous figures as shown in Figure~\ref{fig:optical_illusions}.

Depending on the context, i.e. if it is Easter or Christmas~\cite{Brugger1993}, Figure~\ref{fig:duck_rabbit} can be either a duck or a rabbit.
Likewise, influenced by own-age social biases~\cite{Nicholls2018}, Figure~\ref{fig:young_old_women} can be either a young lady or an old woman.
Humans categorize images based on various types of context.
Known categories are based on visual features or semantic concepts~\cite{Biederman1987RecognitionbycomponentsAT}, but may also be based on other information such as attributes describing their function.
Accordingly, neuroscience has shown that the human brain encodes visual input into individual contextual object representations~\cite{DiCarlo2007UntanglingIO,Greene2020DisentanglingTI,Wardle2020RecentAI}, namely visual, taxonomical, and functional~\cite{Martin2016GRAPESGroundingRI}.
Concretely, in a visual context, images of a drum and a barrel have a high similarity, as they share similar visual features.
In a taxonomical context, a drum would be similar to a violin, as they both are musical instruments.
And in a functional context, the drum would be similar to a hammer, since the same action of hitting can be performed with both objects~\cite{Bracci2017}.

Whereas there is much evidence that intelligent machines should also represent information in contextualized embeddings, deep neural networks (DNNs) form their object representations based only on the feature distribution of the image dataset~\cite{DBLP:conf/iclr/BrendelB19,DBLP:conf/cvpr/ZhangTK20}.
Therefore, they fail if the objects are placed in an incongruent context that was not present in previous seen images~\cite{DBLP:conf/eccv/BeeryHP18}.

For the scope of this work we investigate the following research questions:
\begin{itemize}
    \item \textbf{RQ1} - Can context provided in form of a KG influence learning image representations of a DNN, the final accuracy, and the image predictions?
    \item \textbf{RQ2} - Can context help to avoid critical errors in domain changing scenarios where DNNs fail?
\end{itemize}

To enable standard DNNs to build contextual object representations, we provide the context using a knowledge graph (KG) and its corresponding knowledge graph embedding ($\vec{h}_{KG}$).
Similar to the process in the human brain, we conduct experiments with three different types of contexts, namely visual context, taxonomical context, and functional context~\ref{fig:types_of_context}.
We provide two versions of knowledge infusion into a DNN and compare the induction of different contextual models in depth by quantitatively investigating their learned contextual embedding spaces using class-related cosine similarities.
In addition we evaluate our approach quantitatively by comparing their final accuracy on object recognition tasks on source and target domains and provide insights and challenges.
The structure of this paper is organized as follows:
Section~\ref{sec:related-work} outlines related work.
In Section~\ref{ssec:Contextual View Extraction} we introduce the three different types of context and an option to model these views in a contextual knowledge graph.
Section~\ref{sec:Learning Contextual Image Representations} shows two ways of infusing context into a visual DNN.
In Section~\ref{sec:Experiments} we conduct experiments on seven image datasets in two transfer learning scenarios.
In Section~\ref{sec:Discussion and Insights} we answer the research questions and summarize the main insights of our approach.

\section{Preliminaries}
\label{sec:preliminaries}

\paragraph{Contextual Image Representations in the Brain.}
Cognitive and neuroscience research has recently begun to investigate the relationship between viewed objects and the corresponding fMRI scan activities of the human brain.
It is assumed that the primate visual system is organized into two separate processing pathways in the visual cortex, namely, the \emph{dorsal pathway} and the \emph{ventral pathway}.
While the dorsal pathway is responsible for the spatial recognition of objects as well as actions and manipulations such as grasping, the ventral pathway is responsible for recognizing the type of object based on its form or motion~\cite{Xi2022}.
Bonner et al.~\cite{Bonner2021} recently showed that the sensory coding of objects in the ventral cortex of the human brain is related to statistical embeddings of object or word co-occurrences.
Moreover, these object representations potentially reflect a number of different properties, which together are considered to form an object concept~\cite{Martin2016GRAPESGroundingRI}.
It can be learned based on the context in which the object is seen.
For example, an object concept may include the visual features, its taxonomy, or the function of the object~\cite{Wardle2020RecentAI,Greene2020DisentanglingTI}.

\paragraph{Image Representations in the DNN.}
Recent work has shown that while the performance of humans, monkeys, and DNNs is quite similar for object-level confusions, the image-level performance does not match between different domains~\cite{Wardle2020RecentAI}.
In contrast to visual object representations in the brain, which also include high level contextual knowledge of concepts and their functions, image representations of DNNs only depend on the statistical co-occurrence of visual features and a specific task.
We consider the context extracted from the dataset as dataset bias.
Even in balanced datasets, i.e., datasets containing the same number of images for each class, there still exists imbalance due to overlap of features between different classes.
For instance, it must be taken into account that a cat and a dog have similar visual features and that in composite datasets certain classes can have different meta-information for the images, such as illumination, perspective or sensor resolution.
This dataset bias leads to predefined neighborhoods in the visual embedding space, as well as predefined similarities between distinct classes.
In a DNN, an \emph{encoder network} $E(\cdot)$ maps images $\vec{x}$ to a visual embedding $\vec{h}_{v} = E(\vec{x}) \in \mathbb{R}^{d_E}$, where the activations of the final pooling layer and thus the representation layer have a dimensionality $d_E$, where $d_E$ depends on the encoder network itself.

\paragraph{Contextual Representations in the KG.}
A knowledge graph is a graph of data aiming to accumulate and convey real-world knowledge, where entities are represented by nodes and relationships between entities are represented by edges~\cite{DBLP:journals/corr/abs-2003-02320}.
We define a \emph{generic knowledge graph} ($GKG$) as a graph of data that relates different classes of a dataset based on defined contextual properties.
These contextual properties can be both learned and manually curated.
They bring in prior knowledge about classes, even those that may not necessarily be present in the image dataset, and thus place them in contextual relationships with each other.
A KG comprises a set of triples $G = {H, R, T}$, where $H$ represents entities, $T \subseteq E \times L $ denotes entities or literal values and $R$, is a set of relationships connecting $H$ and $T$.

\section{Learning Contextual Image Representations}
\label{sec:Learning Contextual Image Representations}

\begin{figure}[tb]
    \centering
    \includegraphics[width=\textwidth]{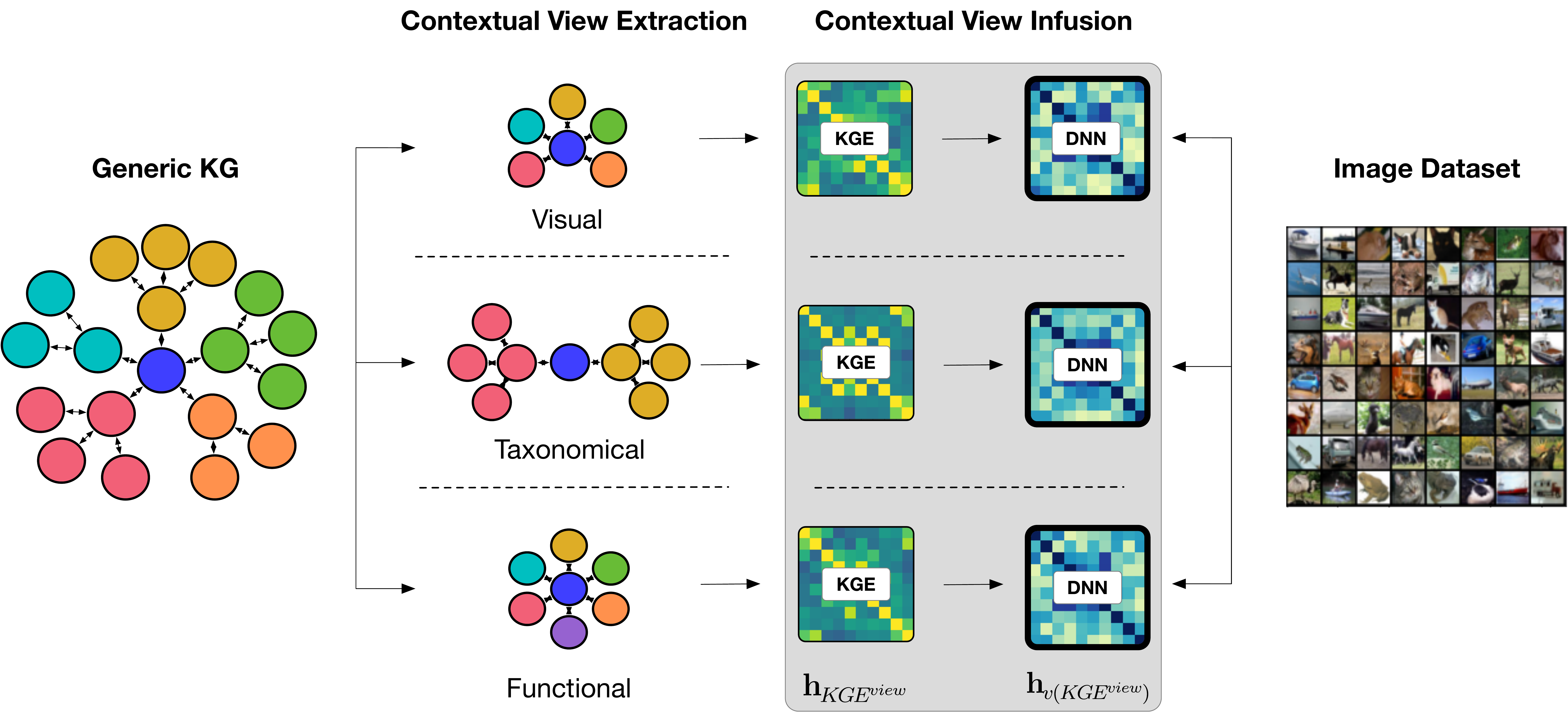}
    \caption{Our approach to learn contextual image representations consists of two main parts: 1) the \emph{contextual view extraction}; and 2) the \emph{contextual view infusion}.}
    \label{fig:Context Framework}
\end{figure}

The framework, as shown in Figure~\ref{fig:Context Framework} consists of two main parts: 1) the \emph{contextual view extraction}, where task relevant knowledge is extracted from a generic knowledge graph; and 2) the \emph{contextual view infusion}, where the contextual view is infused into the DNN.

\subsection{Contextual View Extraction}
\label{ssec:Contextual View Extraction}

A knowledge graph can represent prior knowledge encoded with rich semantics in a graph structure.
A $GKG$ encapsulating $n$ contextual views:
$$ GKG \supseteq \{GKG^1, GKG^2, ..., GKG^n\}$$
is a collection of heterogeneous knowledge sources, where each contextual view defines specific relationships between encoded classes.
However, for a particular task only a specific part of a $GKG$ can be relevant.
Thus, a subgraph containing a single contextual view:
$$ GKG^{view} = query(GKG; view) $$
or a combination of views is extracted from a $GKG$.
Since object recognition models are deployed in the real world that differs from their training domain, it is necessary to encode prior knowledge that is not present in the dataset.

\begin{figure}[tb]
    \centering
    \includegraphics[width=\textwidth]{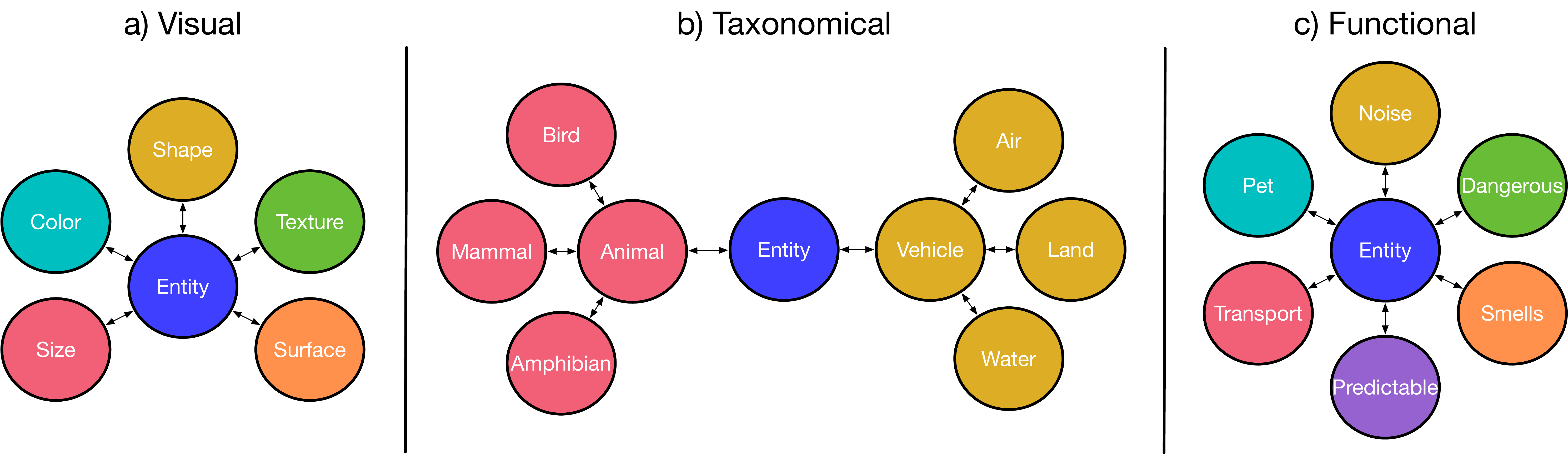}
    \caption{Context can occur in various ways. 
    Aligned to insights of how humans perceive the world, we present three contextual views of a generic knowledge graph, namely the visual, taxonomical, and functional view.}
    \label{fig:types_of_context}
\end{figure}

Based on image representations in our brain and on how humans tend to classify objects, we introduce three distinct types of contextual views as shown in Figure~\ref{fig:types_of_context}.
The first contextual view is based on visual, the second view is based on taxonomical, and the third view is based on functional properties.

\paragraph{Visual Context.}
\label{ssec:Visual Context}
The visual view ($GKG^{v}$) describes high-level visual properties of the classes, for instance properties describing color, shape, or texture.
These properties may or may not be present in the image data set.
For example if all horses in the dataset are white, we want to encode that horses can also occur in different colors.

\paragraph{Taxonomical Context.}
\label{ssec:Taxonomical Context}
The taxonomical view ($GKG^{t}$) describes class relationships based on hierarchical schemes.
A taxonomy is built by experts and can contain categories based on concepts from biology, living place, feeding method, etc. 
For instance, a biological taxonomy separate animals from vehicles and divides them into further subcategories.

\paragraph{Functional Context.}
\label{ssec:Functional Context}
The functional view ($GKG^{f}$) contains properties describing the function of a class.
It is known that tools are categorized in the human brain based on their function~\cite{Martin2016GRAPESGroundingRI}.
In that sense properties as hit, rub, or drill would determine the category of a given tool.
However, to broaden the scope, additional functional properties such as noise, transport, or smell can be introduced.

\subsection{Contextual View Infusion}
\label{ssec:Contextual View Infusion}

\begin{figure}[tb]
    \centering
    \includegraphics[width=\textwidth]{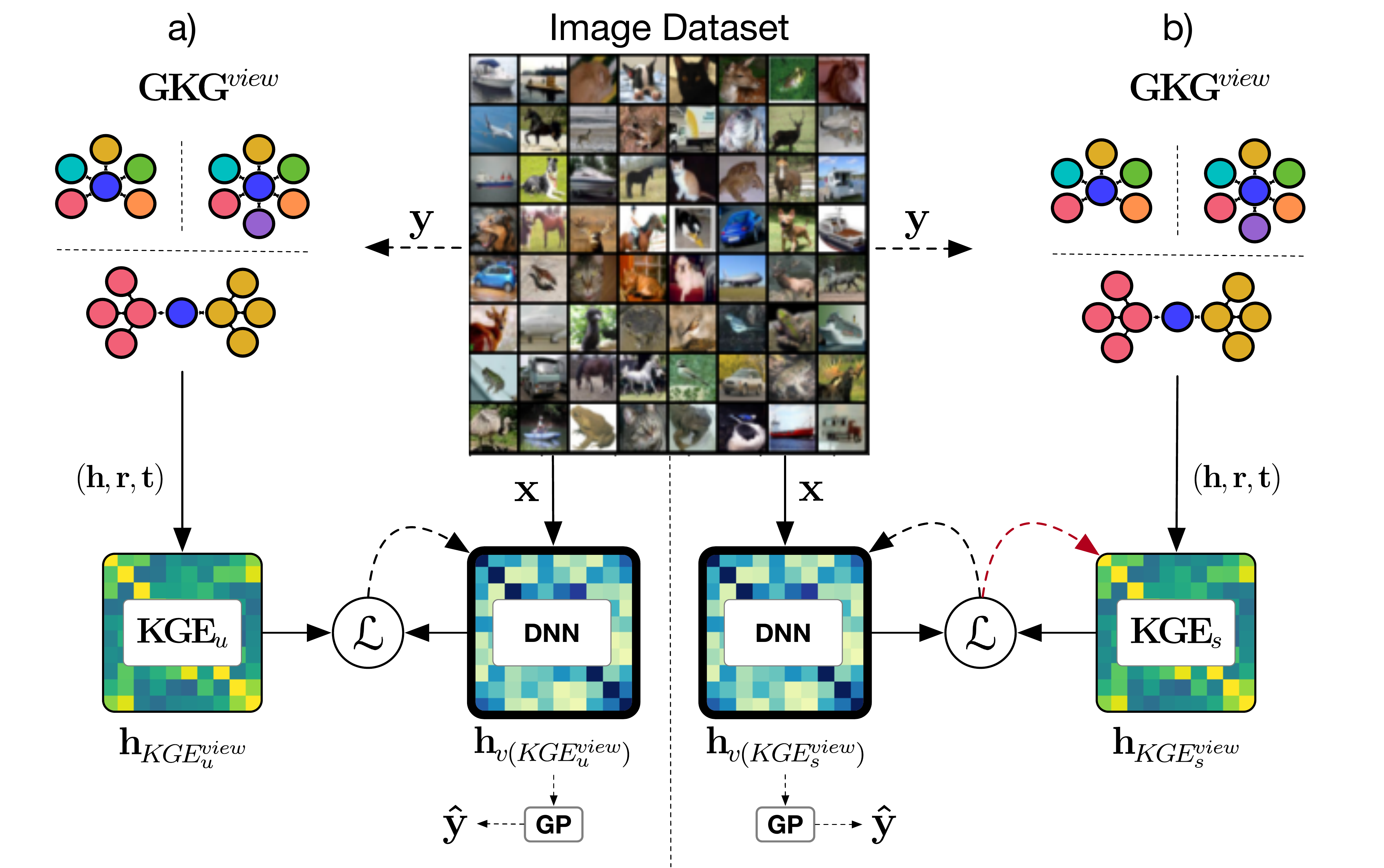}
    \caption{\textbf{Contextual view infusion}.
    The contextual object recognition model (DNN) is trained in two different ways:
    a) using the KG as a trainer, where $KGE_u$ uses no supervision of the image data; or b) using the KG as a peer, where $KGE_s$ uses supervision of the image data.
    Images $\vec{x}$ are fed into the DNN, producing $\vec{h}_{v({KGE}^{view})}$ which is compared with $\vec{h}_{{KGE}^{view}}$ using the KG-based contrastive loss.
    In a second step, a gaussian process (GP) or linear layer is trained to predict the class labels $\vec{y}$ of $\vec{x}$ based on the trained $\vec{h}_{v({KGE}^{view})}$.}
    \label{fig:Network Pipeline}
\end{figure}

When transferring the knowledge from the $GKG^{view}$ using a knowledge graph embedding method ($KGE$) into a knowledge graph embedding:
$$ \vec{h}_{KGE^{view}} = KGE(GKG^{view}) $$
graph based relationships are transferred into spatial relationships.
Intuitively, a different context leads to a different representation in the vector space, where $\vec{h}_{{KGE}^{view}}$ reflects all relationships that are modelled in $GKG^{view}$.

As illustrated in Figure~\ref{fig:Network Pipeline}, we present two different ways of learning a visual context embedding $\vec{h}_{v(GKG^{view})}$ following Monka et al.~\cite{DBLP:journals/semweb/MonkaHR22}.
The first one is $DNN_{{KGE}_u^{view}}$, which uses the knowledge graph as a trainer~\cite{DBLP:conf/semweb/MonkaH0R21} and thus learns $\vec{h}_{KGE_u^{view}}$ without any supervision of image data.
The second version is $DNN_{{KGE}_s^{view}}$, which uses the knowledge graph as a peer and thus learns $\vec{h}_{v(KGE_s^{view})}$ and $\vec{h}_{KGE_s^{view}}$ jointly with additional supervision of image data.

Both versions use the contrastive loss to align the image embedding $\vec{h}_{v(KGE^{view})}$ of the images $\vec{x}$ and the DNN with the knowledge graph embedding $\vec{h}_{KGE^{view}}$ of the label information.
A batch consists of N augmented training samples.
The KG-based contrastive loss is constructed using the individual anchor losses as given by:
$$
    \mathcal{L}_{{KGE}^{view}} = \sum^{N}_{i=1} \mathcal{L}_{{KGE}^{view},i}.
$$

Within a batch, an anchor image $i$ $\in \{1...2N\}$ is selected that corresponds to a specific class label $\vec{y}_i$, where $\vec{y}_i$ points to its knowledge graph embedding $\vec{h}_{{KGE}^{view},i}$.
Positive images $j$ are all images of the batch that correspond to the same class label as the anchor $i$.
The numerator in the loss function computes a similarity score between $\vec{h}_{{KGE}^{view},i}$ and the image embeddings $\vec{h}_{v({KGE}^{view}),j}$.
The denominator computes the similarity score between $\vec{h}_{{KGE}^{view},i}$ and the image embeddings $\vec{h}_{v(KGE),k}$ of all images of the other classes in the batch.
As a similarity score, we choose the cosine similarity, which however can be replaced by others.
$\mathds{1}_{k \neq i} \in \{0, 1\}$ is an indicator function that returns 1 iff $k \neq i$ evaluates as true, and $\tau > 0$ is a predefined scalar temperature parameter.

\[ 
    \centering
    \resizebox{\textwidth}{!}{$\mathcal{L}_{KGE^{view},i}
    =
    \frac{-1}{2N_{\vec{y}_i} - 1} \sum_{j = 1}^{2N} \mathds{1}_{i \neq j} \cdot \mathds{1}_{\vec{y_i=\vec{y}_j}} \cdot \log \frac{\exp{(\vec{h}_{KGE^{view},i} \cdot \vec{h}_{v(KGE^{view}),j} / \tau)}}{\sum^{2N}_{k=1} \mathds{1}_{i \neq k} \exp{(\vec{h}_{KGE^{view},i} \cdot \vec{h}_{v(KGE^{view}),k} / \tau)}}
    $}
\]

\paragraph{Prediction.}
To predict the class labels of unknown images it is common to train a linear layer ($LL$) or to use a gaussian process ($GP$) on top of $\vec{h}_{v(KGE^{view})}$.
For $GP$, we run the whole training dataset through the trained DNN and calculate the mean and covariance matrices for all the classes in $\vec{h}_{v(KGE^{view})}$.
$GP$ and $LL$, both calculate decision boundaries in $\vec{h}_{v(KGE^{view})}$ for all the classes of the dataset.
At inference, where the goal is to predict the class label of an unknown image, $GP$ or $LL$ assign probabilities if an image belongs to a specific class.
The maximal probability is chosen to be the final prediction.

\section{Experiments}
\label{sec:Experiments}
The goal of our empirical investigations is to provide an answer to \textbf{RQ1} and \textbf{RQ2}.
Therefore we conduct experiments with seven datasets in the two specific domain generalization settings, Cifar10 and Mini-ImageNet.
For both experiments, we build separate GKGs that include three different contextual views, the visual ($GKG^{v}$), the taxonomical ($GKG^{t}$), and the functional ($GKG^{f}$) view, respectively.
Based on the framework in Section~\ref{sec:Learning Contextual Image Representations}, we use ${GKG}^{view}$ to learn a contextual DNN in combination with image data.
We evaluate and compare both versions of our approach, $DNN_{{KGE}_u^{view}}$ and $DNN_{{KGE}_s^{view}}$.

\subsection{Implementation details}
\label{ssec:Implementation details}
For both experiments, we use a similar implementation of our approach.
From the $GKG$, we extract various $GKG^{view}$s using respective SPARQL queries.
A ResNet-18 architecture is used as a DNN-backend, with a 128-dimensional MLP as the head.
We train all configurations using an ADAM optimizer, a learning rate of 0.001, no weight decay, and a cosine annealing scheduler with a learning decay rate of 0.1. 
The images are augmented via random cropping, random horizontal flipping, color jittering, random grayscaling, and resizing to 32x32 pixels.
All models are trained for 500 epochs.
For a) $DNN_{{KGE_u}^{view}}$ we transform $GKG^{view}$ into vector space using a graph auto encoder (GAE)~\cite{DBLP:journals/corr/KipfW16a}, which we denote as the $DNN_{{GAE}^{view}}$ model.
Our GAE comprises two convolutional layers, with a hidden layer dimension of 128.
We train the GAE using an ADAM optimizer with a learning rate of 0.01 for 500 epochs.
For b) $DNN_{{KGE}_s^{view}}$, a graph attention network (GAT)~\cite{DBLP:conf/iclr/VelickovicCCRLB18} is trained in combination with the image data, denoted as the $DNN_{{GAT}^{view}}$ model.
The GAT consists of two GAT-layers with 256 hidden dimensions, 8 heads, and an output dimension of 128.
Training is performed via the same KG-based contrastive loss from the images in addition to the $GKG^{view}$ input.
We optimize the GAT using an ADAM optimizer with a learning rate of 0.001 and no weight decay.

\subsection{Experiments on Cifar10}
\label{ssec:Experiments on Cifar10}

\paragraph{Dataset settings.}
The source domain Cifar10~\cite{cifar10dataset} consists of 6000 32x32 color images for each of the 10 classes, namely airplane, bird, automobile, cat, deer, dog, horse, frog, ship, and truck.
The target domain Stl10~\cite{stl10dataset} includes 500 96x96 color images for each of the 10 classes, namely airplane, bird, automobile, cat, deer, dog, horse, monkey, ship, and truck.

\paragraph{Knowledge graph construction.}
We build a $GKG$ that includes the previously discussed three types of context, as shown in Figure~\ref{fig:types_of_context}.
$GKG^{v}$ contains visual properties like: 
\emph{hasBackground}: air, forest, water; 
\emph{hasColor}: black, blue, brown; 
\emph{hasPart}: eyes, legs, wings; 
\emph{hasShape}: rectangular, ellipsoid, cross; 
\emph{hasSize}: large, medium, small;
or \emph{hasTexture}: dotted, striped, uniform.
$GKG^{t}$ contains a taxonomy of the classes using the type-relation.
For example, the class \emph{Horse} is-a \emph{Mammal} and is-an \emph{Animal} or the class \emph{Ship} is-a \emph{Water-vehicle} and is-a \emph{Vehicle}.
$GKG^{f}$ defines the function of the class, e.g. properties like: 
\emph{hasMovement}: drive, fly, swim; 
\emph{hasSound}: bark, meow, vroom; 
\emph{hasSpeed}: fast, medium, slow; 
\emph{hasWeight}: heavy, light, middle.
Our GKG contains in total 34 classes, 16 object properties, and 65 individuals.
Please note that our $GKG$ is only an example and we are aware that there are unlimited possibilities of how and what type of knowledge can be modeled in a knowledge graph.

\paragraph{Evaluation}
\begin{figure}[tb]
    \centering
    \includegraphics[width=\textwidth]{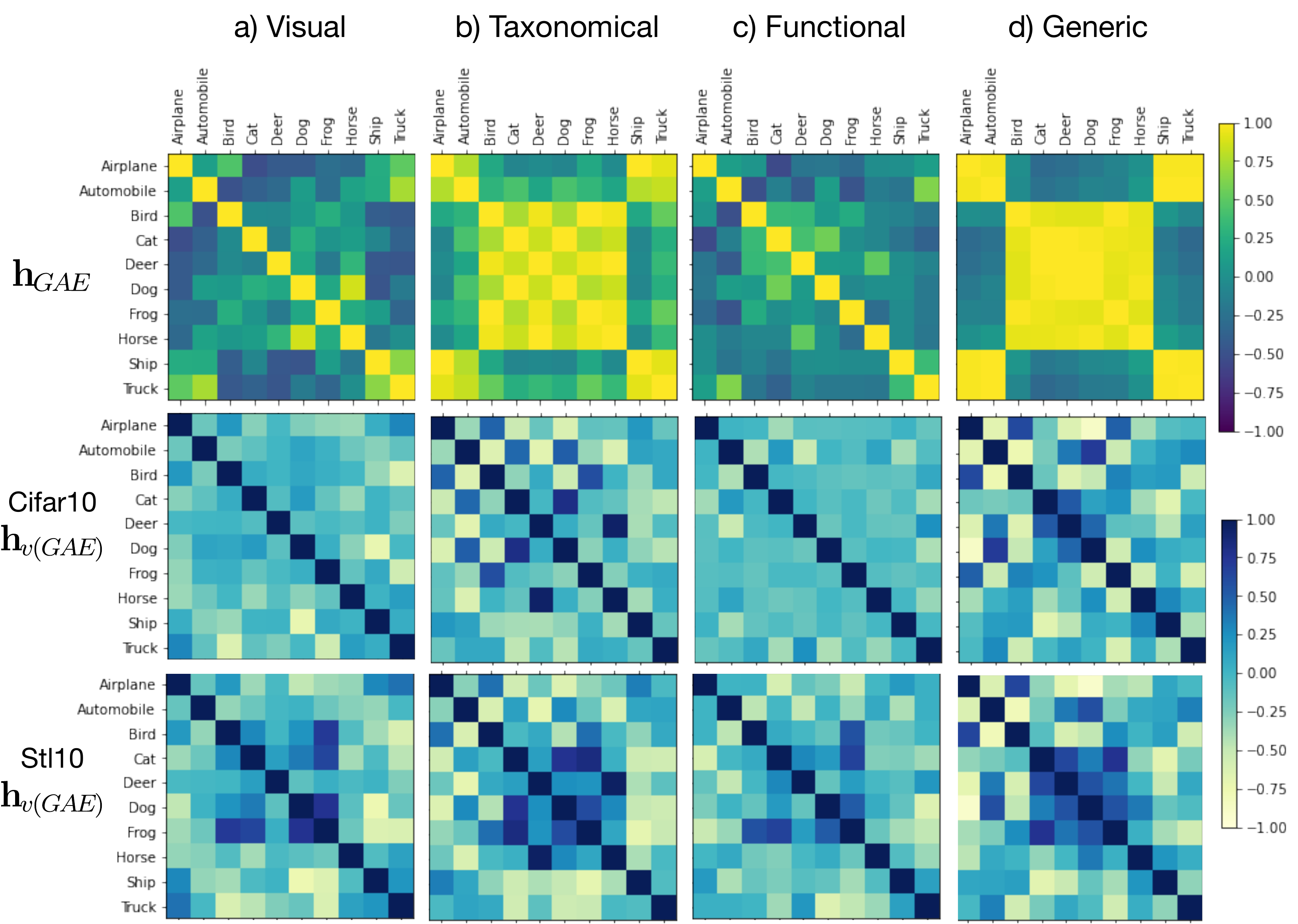}
    \caption{We compare $\vec{h}_{GAE^{view}}$ and $\vec{h}_{v(GAE^{view})}$ based on: a) the visual view; b) the taxonomical view; c) the functional view; and d) the full generic KG.
    To investigate how the semantic relationships are reflected in the embeddings, we illustrate the individual cosine similarities between the classes of the Cifar10 and the Stl10 dataset.}
    \label{fig:GAE Context}
\end{figure}

To evaluate our approach we first investigate the learned embeddings, if and how semantic relationships from $GKG^{view}$ are reflected in $\vec{h}_{GKG^{view}}$. 
Second, we compare the individual class accuracies to see how these relationships influence the final object recognition.
Figure~\ref{fig:GAE Context} shows an analysis: a) the visual view; b) the taxonomical view; and c) the functional view.
For every cell in $\vec{h}_{GAE^{view}}$ we calculate the cosine similarity between the corresponding nodes, i.e. the classes of the image dataset, and for $\vec{h}_{v(GAE^{view})}$ we calculate the class-means of the image representations.
Since the goal is to learn contextual image classifiers, we investigate if context is transferred to $\vec{h}_{GKG^{view}}$ and $\vec{h}_{v(GKG^{view})}$, respectively.
It can be seen that semantic relationships provided by the $GKG^{view}$ are reflected in $\vec{h}_{GAE^{view}}$.
In $\vec{h}_{GAE^{v}}$, the airplane has the highest similarity to the truck and the bird,
in $\vec{h}_{GAE^{t}}$, the airplane has the highest similarity to the ship,
in $\vec{h}_{GAE^{f}}$, the airplane has the highest similarity to the automobile, and $\vec{h}_{GAE}$ the airplane has a high similarity to all vehicles.
Further, one notices that taxonomical and generic $\vec{h}_{GAE}$ have two main distinctive groups in the embedding space.
In $\vec{h}_{{GAE}^t}$ and $\vec{h}_{{GAE}}$ vehicles and animals have a high inter-cluster, but a small intra-cluster variance.
For $\vec{h}_{v(GAE^{view})}$, we observe that similarities in the $GKG^{view}$ and $\vec{h}_{GAE^{view}}$ are only partially reflected.
All $\vec{h}_{v(GAE^{view})}$ seem to have a similar underlying pattern of the class distribution, with minor differences.
We think that implicit relations between class features interfere with the similarities given by $\vec{h}_{GAE}$ and the $GKG$.
Further we retrieve different distributions for either Cifar10 or Stl10.
This behaviour can be explained by the distribution shift between source and target domain.
While the network attempts to separate classes in the training domain Cifar10, this separation is less successful in the testing domain Stl10.

\begin{table}[tb]
    \centering
    \begin{subtable}{\linewidth}
    \centering
    \subcaption{Results on Cifar10}
    \begin{tabular}{lrrrrrrrrrrr}
    \toprule
    {Cifar10} &  Airplane &  Auto &  Bird &   Cat &  Deer &   Dog &  Frog &  Horse &  Ship &  Truck &   All \\
    \midrule
    SupSSL             &      95.1 &        97.0 &  91.8 &  83.9 &  92.9 &  85.7 &  96.0 &   93.5 &  96.8 &   95.9 &  92.9 \\
                      &        &          &    &    &    &    &    &     &    &     &    \\
    $\Delta DNN_{{GAE}^{v}}$                  &      -1.2 &         \textbf{0.5} &  -2.6 &  -0.2 &   \textbf{2.3} &  -0.8 &  -0.2 &   -1.1 &  -0.5 &   -0.9 &  -0.5 \\
    $\Delta DNN_{{GAE}^{t}}$                  &      -0.9 &        -0.6 &  -1.2 & -30.8 & -29.8 &  -0.2 &  -2.2 &   -1.6 &  -1.5 &   -1.3 &  -7.0 \\
    $\Delta DNN_{{GAE}^{f}}$                  &       \textbf{1.0} &         0.2 &  -1.1 &   \textbf{1.9} &   0.1 &   0.6 &   \textbf{0.7} &    1.2 &  -0.1 &   -0.4 &   \textbf{0.4} \\
    $\Delta DNN_{GAE}$                         &      -0.7 &         0.0 &  -2.3 &   0.4 &   0.6 &  -0.6 &  -1.1 &    0.0 &   0.3 &   -1.8 &  -0.5 \\
    \arrayrulecolor{black!30}\midrule\arrayrulecolor{black}
    $\Delta DNN_{{GAT}^{v}}$                  &      -0.6 &        -0.3 &   \textbf{0.2} &   0.3 &   0.1 &  -1.0 &   0.3 &    0.9 &   \textbf{0.7} &   -0.8 &  -0.0 \\
    $\Delta DNN_{{GAT}^{t}}$                  &      -0.9 &         0.0 &  -1.7 &   1.8 &   0.1 &   1.0 &   0.4 &    0.5 &  -0.1 &    0.3 &   0.1 \\
    $\Delta DNN_{{GAT}^{f}}$                  &      -0.4 &         \textbf{0.5} &  -3.0 &   1.7 &   1.5 &  -0.4 &   0.4 &    0.8 &   0.4 &   -0.1 &   0.1 \\
    $\Delta DNN_{GAT}$                          &      -1.0 &         0.3 &  -1.8 &   1.2 &  -0.3 &   \textbf{2.0} &  -0.5 &    \textbf{1.7} &   0.0 &    \textbf{0.7} &   0.2 \\
    \bottomrule
    \end{tabular}
    \end{subtable}
    \begin{subtable}{\linewidth}
    \centering
    \subcaption{Results on Stl10}
    \label{tab:Results on Stl10}
    \begin{tabular}{lrrrrrrrrrrr}
    \toprule
    {Stl10} &  Airplane &  Auto &  Bird &   Cat &  Deer &   Dog &  Frog &  Horse &  Ship &  Truck &   All \\
    \midrule
    SupSSL                      &      85.4 &        86.9 &  82.4 &  56.6 &  91.5 &  60.5 &   - &   76.5 &  84.5 &   74.1 &  77.6 \\
                 &        &          &    &    &    &    &    &     &    &     &    \\
    $\Delta DNN_{{GAE}^{v}}$          &       1.0 &         0.2 &  -2.6 &   \textbf{3.4} &   \textbf{1.2} &  -4.5 &   - &   -4.8 &  -0.6 &    3.9 &  -0.3 \\
    $\Delta DNN_{{GAE}^{t}}$    &       2.4 &        -1.0 &  -1.5 & -10.1 & -32.9 &   0.2 &   - &   -0.5 &  -1.6 &   -1.6 &  -5.2 \\
    $\Delta DNN_{{GAE}^{f}}$     &       1.9 &        -0.8 &  -1.3 &   1.4 &  -2.4 &  -0.5 &   - &    \textbf{3.4} &   0.9 &    3.3 &   \textbf{0.7} \\
    $\Delta DNN_{GAE}$ &       0.4 &         \textbf{0.5} &  -1.9 &   1.8 &  -1.5 &   2.6 &   - &   -1.4 &  -0.6 &    2.1 &   0.2 \\
    \arrayrulecolor{black!30}\midrule\arrayrulecolor{black}
    $\Delta DNN_{{GAT}^{v}}$    &       0.5 &        -0.9 &   \textbf{2.6} &  -0.6 &  -0.1 &   0.5 &   - &    0.5 &   1.0 &    0.0 &   0.4 \\
    $\Delta DNN_{{GAT}^{t}}$ &       1.0 &        -2.1 &  -0.5 &   1.9 &  -0.4 &   0.8 &   - &    0.5 &   \textbf{1.6} &    3.0 &   0.6 \\
    $\Delta DNN_{{GAT}^{f}}$  &       \textbf{2.7} &        -0.3 &  -1.5 &  -0.7 &  -1.0 &  -2.6 &   - &    0.0 &   0.4 &    1.8 &  -0.1 \\
    $\Delta DNN_{GAT}$   &      -1.6 &        -1.0 &  -2.6 &  -2.2 &  -1.2 &   \textbf{2.8} &   - &    3.1 &   1.2 &    \textbf{4.3} &   0.3 \\
    \bottomrule
    \end{tabular}
    \end{subtable}
    \caption{Comparison of the individual class accuracies for the Cifar10 dataset as training domain and the Stl10 dataset as testing domain.
    We compare the contextual view trained DNNs against their baseline SupSSL.}
    \label{tab:Accuracy Cifar10}
\end{table}

In Table~\ref{tab:Accuracy Cifar10} we compare the final object recognition accuracy of the contextual DNNs, compared to their baseline SupSSL.
SupSSL is the same model trained with the supervised contrastive loss~\cite{DBLP:conf/nips/KhoslaTWSTIMLK20} and without auxiliary context.
We observe that for different contextual infusions the overall accuracy is not significantly impacted.
For Cifar10 $\Delta DNN_{{GAE}^t}$ with $-7.0$ is the worst performing model, whereas $\Delta DNN_{{GAE}^f}$ with $0.4$ is the best performing model.
We marked the best performing model for every class in bold.
It can be seen that for every class a different contextual model is outperforming the others.
It also shows that context influences the focus a DNN puts on predicting a specific class.
Table~\ref{tab:Results on Stl10} shows the relative accuracies of the contextual models on the Stl10 dataset.
Note that the models are only trained on Cifar10 data.
The goal of that domain generalization scenario is to test the robustness of the models.
When evaluated on the target domain, it can be observed that almost in every contextual model the relative accuracy is increased compared to the baseline with no contextual knowledge.
In scenarios where the domain changes, we observe strange phenomena occurring such that the model with the second worst performance $DNN_{GAE}^{t}$ for the class Aircraft of the Cifar10 dataset is the model with the second best performance for Aircraft on Stl10.
However, for most of the classes, we see a trend that the best performing model for a class in Cifar10 tends to perform also better on the target domain.

\subsection{Experiments on Mini-ImageNet}
\label{ssec:Experiments on Mini-ImageNet}

\paragraph{Dataset settings.}
We use Mini-ImageNet, a subset of the ImageNet dataset, as our training domain.
It contains 100 classes, each having 600 images of size 84 × 84. 
As testing domain we use ImageNetV2~\cite{DBLP:conf/icml/RechtRSS19} comprising 10 new test images per class, ImageNet-Sketch~\cite{wang2019learning} with 50 images per class, ImageNet-R~\cite{hendrycks2020many}, which has 150 images in the style of art, cartoons, deviantart, and ImageNet-A~\cite{hendrycks2019nae} with 7.500 unmodified real-world examples.

\paragraph{Knowledge graph construction.}
Our $GKG$ is build using the three contextual views as depicted in Figure~\ref{fig:types_of_context}.
$GKG^{v}$ contains visual properties, e.g. \emph{hasColor}: black, blue, brown; \emph{hasTexture}: dotted, striped, uniform; \emph{hasSize}: large, medium-large, small; and \emph{hasShape}: ellipsoid, quadratic, rectangular.
$GKG^{t}$ contains a taxonomy of the classes using the type-relation.
Following DBpedia~\cite{DBLP:conf/semweb/AuerBKLCI07}, the class \emph{Malamute} is-a \emph{Dog}, is-a \emph{Mammal}, is-an \emph{Animal}, is-an \emph{Eukaryote}, and is-a \emph{Species}.
$GKG^{f}$ defines the function of a class with properties like: \emph{hasSpeed}: fast, static, slow; \emph{hasWeight}: heavy, light, middle; or \emph{hasTransportation}: goods, none, people.
Our GKG contains in total 166 classes, 14 object properties, and 183 individuals.

\paragraph{Evaluation.}
\begin{figure}[tb]
    \centering
    \includegraphics[width=\textwidth]{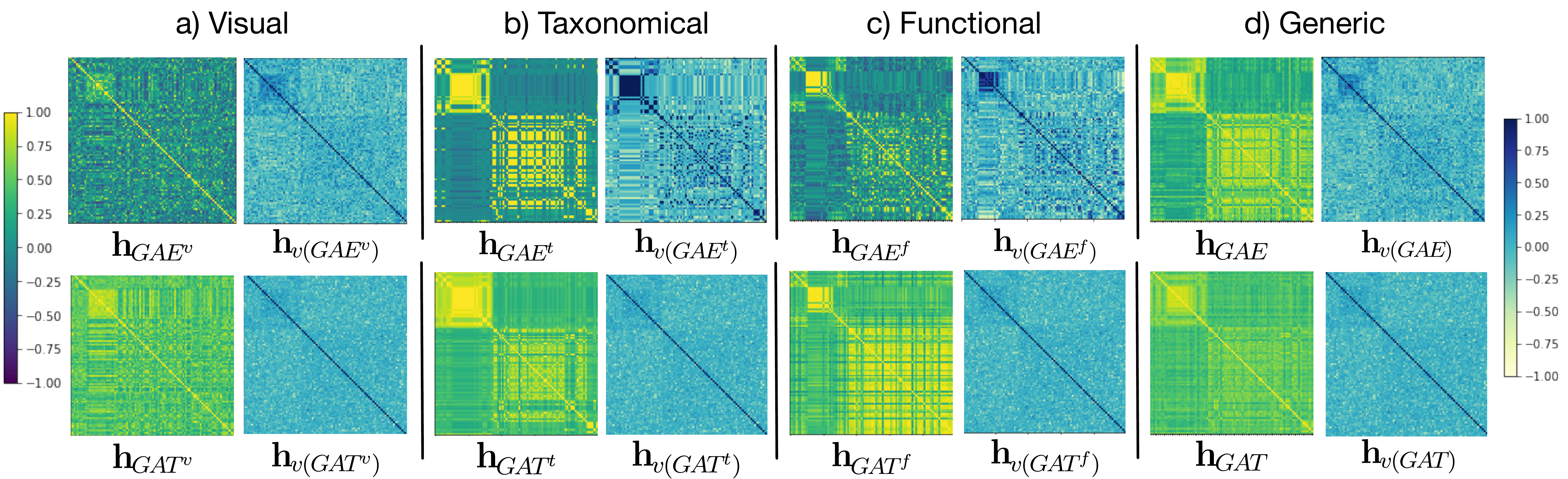}
    \caption{We compare $\vec{h}_{GAE^{view}}$ and $\vec{h}_{v(GAE^{view})}$, as well as $\vec{h}_{GAT^{view}}$ and $\vec{h}_{v(GAT^{view})}$ based on, a) the visual view, b) the taxonomical view, c) the functional view, and d) the full generic KG.
    To investigate how the semantic relationships are reflected in the embeddings, we illustrate the individual cosine similarities between the classes of the Mini-ImageNet dataset.}
    \label{fig:Mini Embeddings}
\end{figure}
Due to the difficulty of deeply investigating 100x100 class similarities, we provide a qualitative overview of the embedding spaces.
Figure~\ref{fig:Mini Embeddings} shows a qualitative comparison of $\vec{h}_{{KGE}^{view}}$ and $\vec{h}_{v({KGE}^{view})}$ of a) the visual view; b) the taxonomical view; c) the functional view; and d) the generic knowledge graph.
Complementing the experiment in Section~\ref{ssec:Experiments on Cifar10}, we illustrate the class similarities of $\vec{h}_{GAT}$ and $\vec{h}_{v({GAT})}$ learned using image data as supervision.
Interestingly, it can be observed that the similarities in $\vec{h}_{GAT}$ and $\vec{h}_{GAE}$ follow a similar pattern, but $\vec{h}_{GAE}$ seems to have a stronger contrast.
However, when investigating the learned image representations in $\vec{h}_{v({GAT})}$ it is hard to spot the differences between the individual contextual models.
\begin{table}[tb]
    \centering
\begin{tabular}{p{3cm}p{1.5cm}p{1.5cm}p{1.5cm}p{1.5cm}p{1.5cm}p{1.5cm}}
\toprule
{ImageNet} &  Mini &  V2 &  Sketch &  R &  A \\
\midrule
SupSSL             &        58.6 &           43.0 &        20.3 &        4.3 &       1.2  \\
\\
$\Delta DNN_{{GAE}^v}$ &        -0.3 &            0.0 &         -0.6 & \textbf{0.2} &        -0.2  \\
$\Delta DNN_{{GAE}^t}$ &       -19.6 &           -13.7 &       -8.8 &        -2.8 &       0.0  \\
$\Delta DNN_{{GAE}^f}$&       -5.2 &           -3.3 &        -2.3 &         -0.7 &       \textbf{0.3}  \\
$\Delta DNN_{{GAE}}$ &         0.8 &            1.6 &        -0.6 &         -0.1 &        -0.1  \\
\arrayrulecolor{black!30}\midrule\arrayrulecolor{black}
$\Delta DNN_{{GAT}^v}$  &        0.9 & \textbf{2.3} &  \textbf{0.2} &         0.0 &        \textbf{0.3}  \\
$\Delta DNN_{{GAT}^t}$   &\textbf{1.3} &           0.6 &        0.1 &         0.1 &        0.0  \\
$\Delta DNN_{{GAT}^f}$  &         0.4 &            0.4 &         0.0 &         -0.1 &       -0.1  \\
$\Delta DNN_{{GAT}}$   &        0.5 &           0.6 &         0.1 &         0.0 &        0.0  \\
\bottomrule
\end{tabular}
    \caption{Comparison of the contextual view models and their SupSSL baseline on the Mini-ImageNet and its derivatives, Mini-ImageNet (Mini), ImageNetV2 (V2), ImageNet-Sketch (Sketch), ImageNet-R (R), and ImageNet-A (A).}
    \vspace{-0.5cm}
    \label{tab:Accuracy Mini-ImageNet}
\end{table}
As depicted in Table~\ref{tab:Accuracy Mini-ImageNet} $DNN_{GAE}^{t}$ and $DNN_{GAE}^{f}$ are outperformed by the baseline $SupSSL$ and the other models with different contextual views by a large margin.
In contrast to the Cifar10 experiment where the least performing model is only $8\%$ worse than the baseline, in Mini-ImageNet the worst is around $34\%$.
Further, we see that $DNN_{{GAT}^{t}}$ does not suffer from constraints given by $GKG^{t}$.
This finding confirms our assumption that a joint training can soften the constraints of the $GKG$.

\begin{figure}[tb]
    \centering
    \includegraphics[width=\textwidth]{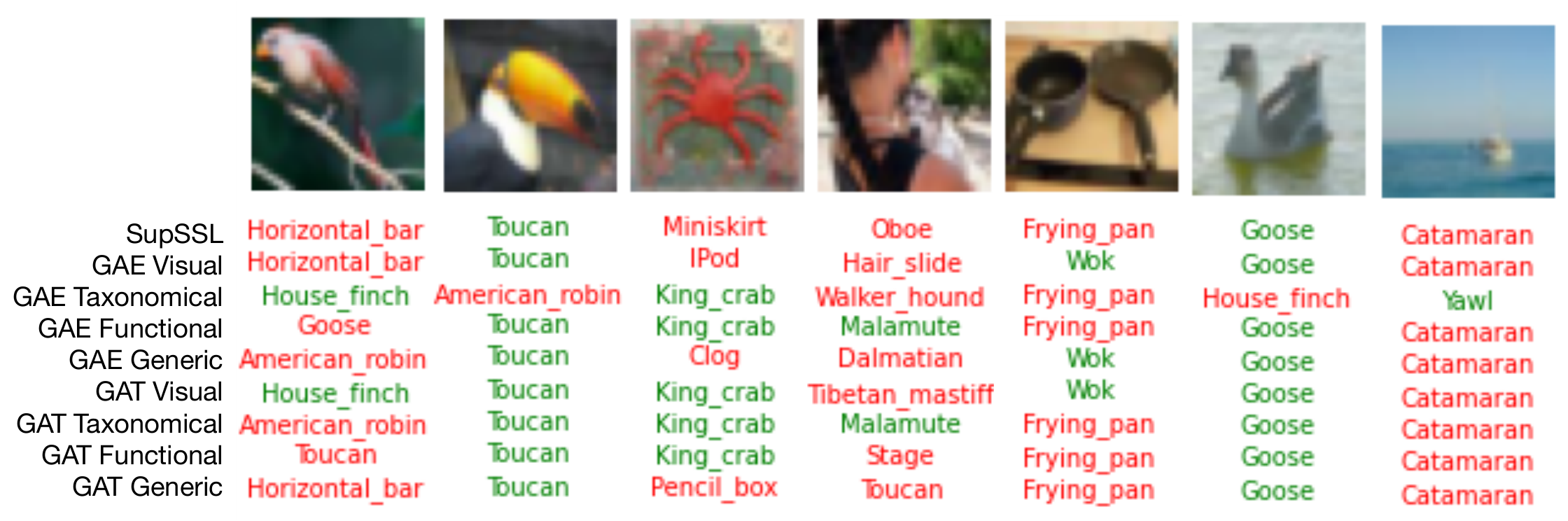}
    \caption{Contextual Predictions of $DNN_{GAE}$ ($GAE$) and $DNN_{GAT}$ ($GAT$) and their contextual view on Mini-ImageNet.
    The contextual view influences the image representation and therefore the final prediction for the same input image.}
    \label{fig:Predictions}
\end{figure}

Similar to the example of ambiguous figures in Figure~\ref{fig:optical_illusions}, our approach enables DNNs to interpret the same image in various ways using contextual views given by a knowledge graph.
The results in Figure~\ref{fig:Predictions} show that for out of distribution images the contextual views play a major role for giving reasonable predictions.
The idea is that some class confusions are not that critical as others.
In that sense, for some tasks it is uncritical to confuse a goose with a house finch as they are both part of the bird family, however confusing a music instrument (oboe), with a dog (malamute) could lead to problems. 
We also see that $DNN_{GAE}$ ($GAE$) and $DNN_{GAT}$ ($GAT$) do not necessarily predict the same image based on the given context.
We believe that further research is needed w.r.t. investigating how to best incorporate context in combination with image data.

\section{Discussion and Insights}
\label{sec:Discussion and Insights}
With our work, we provided a method to infuse context in form of $GKG^{view}$ into DNNs for visual object recognition.
However, knowledge infusion is not straightforward, as problems of machine learning, such as hyper-parameter selection, weight initialization, or dataset dependence, strongly influence the learned representations.
Regarding \textbf{RQ1} - Can context provided in form of a KG influence learning image representations of a DNN, the final accuracy, and the image predictions? - we list the insights obtained from our investigations:\\
\textbf{- $GKG^{view}$ defines class-relationships.}
We showed that various contextual views can be extracted from a $GKG$ and that different views lead to different relationships between classes of the dataset.\\
\textbf{- $\vec{h}_{KGE^{view}}$ needs to reflect $GKG^{view}$.}
The embedding method itself also influences the $\vec{h}_{{KGE}^{view}}$ and the performance of the final prediction model. 
Context can get lost when transferring $GKG^{view}$ into $\vec{h}_{KGE^{view}}$.
Hard constraints either in ${GKG}^{view}$ or produced by the KGE-method, e.g. to represent dissimilar classes in $\vec{h}_{KGE^{view}}$ together, can drastically reduce the prediction accuracy.\\
\textbf{- $\vec{h}_{GAE^{view}}$ is only partially reflected in $\vec{h}_{v(GAE^{view})}$.}
Since data-driven approaches have a strong dependence on the dataset distribution, $\vec{h}_{GAE^{view}}$ only influences $\vec{h}_{v(GAE^{view})}$ to form a hybrid representation.
We see that data augmentation weakens the dataset bias and helps to align $\vec{h}_{v(GAE^{view})}$ with $\vec{h}_{GAE^{view}}$.\\
\textbf{- Joint training reduces the impact of $GKG$.}
Both the learned $\vec{h}_{v({KGE_s})}$ and the achieved accuracy values are only slightly affected by the induced $GKG$.
Neither the qualitative evaluation of $\vec{h}_{v({KGE_s})}$ nor the quantitative evaluation based on accuracy show any significant contextual changes.\\
\textbf{- Context shifts the focus on learning specific classes.}
We assume that the context constraints the DNN and its hypothesis space.
It is known that DNNs tend to memorize spurious correlations that can lead to catastrophic errors in the real world. 
We think that the task of our contextual models is to prevent exactly these errors.
In our experiments, we showed that specific contextual models performed better on specific classes.
We assume that context can shift the overall interest of a DNN to predict a certain class.\\
\textbf{- Context rather influences individual image predictions.}
Similar to the proposed motivation of how humans interpret ambiguous figures we see context influencing the prediction of difficult or undefinable images in the dataset.
Regarding \textbf{RQ2} - Can context help to avoid critical errors in domain changing scenarios where DNNs fail?\\
\textbf{- Context makes more robust against domain changes.}
It can be seen that almost every contextual model increases its relative accuracy compared to the baseline when evaluated on the target domain.
Moreover, contextual models that performed better on the source dataset tend to perform better if domain change occurs.
We argue that $GKG^{view}$ regularizes the strong dependency on the source domain and thus increases the performance on the target domain.


\section{Related Work}
\label{sec:related-work}
Contextual information has always been of great interest for improving computer vision systems.
We structure related work into implicit-contextual visual models, explicit-contextual visual models, and contextual knowledge graph embeddings.


\paragraph{Implicit-contextual visual models} contextualize relationships between visual features that occur in the image itself.
They are used for object priming, where the context defines a prior on the detection parameters~\cite{DBLP:journals/ijcv/Torralba03} or for object detection and segmentation, where boosting is used to relate objects in an image~\cite{DBLP:conf/nips/TorralbaMF04}.
Wu et al.~\cite{DBLP:conf/ciss/WuWK18} improved object recognition by processing object regions and context regions in parallel.
To overcome the drawback of small receptive fields from standard CNNs, extensions that incorporate visual features from far image regions~\cite{DBLP:conf/nips/HuSASV18,DBLP:journals/pami/HuSASW20} or alternative architectures, such as vision transformers (ViTs)~\cite{Yu2021} have been established recently.
Moreover, Gao et al.~\cite{Gao2021} proposed that all modern DNNs are part of the implicit-contextual models since they aggregate contextual information over image regions.

\paragraph{Explicit-contextual visual models} use higher level information like object co-occurrences or semantic concept relationships.
They induce additional contextual information that is either not in the dataset or cannot be automatically extracted by the DNN~\cite{DBLP:conf/iccv/HoiemEH05}.
To create explicit context based on object relations, most methods use scene graphs which describe a scene based on symbolic representations of entities and their spatial and semantic relations.
Scene graphs have been applied to the task of collective or group activity recognition~\cite{DBLP:conf/eccv/ChoiS12,DBLP:conf/cvpr/DengVHM16}, object recognition~\cite{DBLP:journals/corr/abs-1902-00163,DBLP:conf/cvpr/ZhangTK20}, object detection~\cite{DBLP:conf/cvpr/ChenLFG18,DBLP:conf/cvpr/Liu0SC18} and visual question answering~\cite{DBLP:conf/cvpr/TeneyLH17}.
Label graphs~\cite{DBLP:conf/cvpr/HuZDLM16} apply fine-grained labels to an image and are used to improve object recognition and reasoning over object relationships~\cite{DBLP:conf/nips/BattagliaPLRK16}.
Semantic scene graphs extend scene graphs by textual descriptions and fine-grained labels of a scene~\cite{DBLP:conf/cvpr/LiZM20}.
Context-aware zero-shot learning for object recognition~\cite{DBLP:conf/icml/ZablockiBSPG19} or compositional zero-shot learning methods~\cite{DBLP:conf/cvpr/NaeemXTA21} add observed visual primitive states (e.g. old, cute) to objects (e.g. car, dog) to build an embedding space based on visual context.
However, scene correlations need to be addressed very carefully, as implicit-contextual models can heavily depend on learned contextual relationships that are only valid for a specific dataset configuration.
Therefore, work was already done to decorrelate objects and their visual features to improve model generalization~\cite{DBLP:conf/cvpr/SinghMGLFG20}.

\paragraph{Contextual Knowledge Graph Embeddings}
Whereas our approach extracts the contextual views in a previous step before the actual knowledge graph embedding, there exist works that create contextualized KG embeddings based on the full KG.
Werner et al.~\cite{DBLP:conf/esws/WernerRHL21} introduced a KG embedding over temporal contextualized KG facts. 
Their recurrent transformer enables to transform global KGEs into contextual embeddings, given the situation-specific factors of the relation and the subjective history of the entity.
Ning et al.~\cite{DBLP:conf/pakdd/NingQDDZ21} proposed a lightweight framework for the usage of context within standard embedding methods.
Wang et al.~\cite{DBLP:conf/akbc/WangKW20} presented a deep contextualized knowledge graph embedding method that learns representations of entities and relations from constructed contextual entity-relation chains.
Wang et al.~\cite{DBLP:journals/corr/abs-1911-02168} introduced the contextualized KG embedding method (CoKE).
They propose to take the contextual nature of KGs into account, by learning dynamic, flexible, and fully contextualized entity and relation embeddings.

\section{Conclusion and Future Work}
\label{sec:conclusion}
In this work, we proposed a framework for context-driven visual object recognition based on knowledge graphs.
We qualitatively and quantitatively investigated how different contextual views, as well as their embedding and their infusion method, influence the learned DNN.
Further, we have seen that contextual models tend to have a minor impact on the final accuracy, but a major impact on how individual classes or images are represented and predicted.
In particular, for out of distribution data, where data-driven approaches suffer from less knowledge, contextual image representations help to constrain the hypothesis space, leading to more reasonable predictions.
However, there are still challenges to be faced.
We conducted intensive research about a possible context infusion approach and emerging challenges.
On the one hand, we have the implementation of the infusion method, which itself heavily depends on modeling choices, weight initialization, as well as network and hyper-parameter selection.
On the other hand, there is a strong dependence on the image data, which originally comes with an initial dataset bias.
This dataset bias limits the ability to influence image data representations and thus predictions influenced by prior knowledge.
However, our work showed that with deeper investigations of all the influencing parameters knowledge-infused learning is a promising approach to build context-driven and future intelligent systems.

\section{Acknowledgement}

This publication was created as part of the research project "KI Delta Learning" (project number: 19A19013D) funded by the Federal Ministry for Economic Affairs and Energy (BMWi) on the basis of a decision by the German Bundestag.

\bibliographystyle{splncs04}
\bibliography{literature}

\end{document}